\documentclass[10pt,twocolumn,letterpaper]{article}
\usepackage{wacv}
\usepackage{times}
\usepackage{epsfig}
\usepackage{graphicx}
\usepackage{amsmath}
\usepackage{amssymb}
\usepackage{csquotes}
\usepackage{comment}
\usepackage{hyperref}
\usepackage{textcomp}
\usepackage{multirow,fixltx2e}
\usepackage{array}
\newcolumntype{P}[1]{>{\centering\arraybackslash}p{#1}}
\newcolumntype{M}[1]{>{\centering\arraybackslash}m{#1}}
\newcommand{\norm}[1]{\left\lVert#1\right\rVert}
\DeclareMathOperator{\sgn}{sgn}
\usepackage{ragged2e}
\usepackage[linesnumbered,ruled,vlined]{algorithm2e}

\wacvfinalcopy 

\def\wacvPaperID{****} 

\begin{document}
\title{Enforcing Linearity in DNN succours Robustness and Adversarial Image Generation}

\author{\textbf{Anindya Sarkar} \\
Mobiliya\\
{\tt\small anindya.sarkar@mobiliya.com}
\and
\textbf{Nikhil Kumar Gupta} \\
Mobiliya\\
{\tt\small nikhil.gupta@mobiliya.com}
\and
\textbf{Raghu Iyengar} \\
Mobiliya\\
{\tt\small raghu.iyengar@mobiliya.com}
}

\maketitle

\begin{abstract}
   Recent studies on the adversarial vulnerability of neural networks have shown that models trained with the objective of minimizing an upper bound on the worst-case loss over all possible adversarial perturbations improve robustness against adversarial attacks. Beside exploiting adversarial training framework, we show that by enforcing a Deep Neural Network (DNN) to be linear in transformed input and feature space improves robustness significantly. We also demonstrate that by augmenting the objective function with Local Lipschitz regularizer boost robustness of the model further. Our method outperforms most sophisticated adversarial training methods and achieves state of the art adversarial accuracy on MNIST, CIFAR10 and SVHN dataset. In this paper, we also propose a novel adversarial image generation method by leveraging Inverse Representation Learning and Linearity aspect of an adversarially trained deep neural network classifier.

\end{abstract}

\section{Introduction}
\label{sec:1}
Despite the positive outcomes of deep learning \cite{Goodfellow:2016:DL:3086952} in wide ranges of computer vision tasks such as image classification \cite{7780459, Krizhevsky:2012:ICD:2999134.2999257}, object detection \cite{DBLP:journals/corr/GirshickDDM13, DBLP:journals/corr/RedmonDGF15}, semantic segmentation \cite{DBLP:journals/corr/HeGDG17} etc., it is well known that deep neural networks are vulnerable to adversarial attacks. In particular, It has been shown that carefully designed adversarial examples, which are inputs to machine learning models can cause the model to make a mistake \cite{DBLP:journals/corr/CarliniW16a,Goodfellow2014ExplainingAH, DBLP:journals/corr/KurakinGB16, 42503}. The existence of adversarial examples pose a severe security threat to the practical deployment of deep learning models, particularly in safety critical systems, such as autonomous driving \cite{DBLP:journals/corr/abs-1802-06430}, healthcare \cite{inproceedings} etc.

Starting with the seminal work \cite{42503}, there has been extensive work in the area of crafting new adversarial perturbations. These work can be broadly classified in two categories: (i) finding strategies to defend against adversarial inputs \cite{Goodfellow2014ExplainingAH, Tramr2017EnsembleAT, Madry2017TowardsDL}; (ii) new attack methods that are stronger and can break the proposed defense mechanism \cite{Madry2017TowardsDL, DBLP:journals/corr/Moosavi-Dezfooli16}. 

Most adversarial defense strategies can be broken by carefully crafted stronger adversaries as shown in \cite{DBLP:journals/corr/abs-1802-05666, DBLP:journals/corr/abs-1802-00420}. \cite{Madry2017TowardsDL}
proposed robust optimization method, which solves this issue by finding the worst case adversarial samples on the fly during training. Though the resulting models show empirical proofs that they are robust against many standard attacks, \cite{DBLP:journals/corr/abs-1711-07356} observed a phenomenon, which shows \cite{Madry2017TowardsDL} does not always find the worst-case attack and it can not be assured that a different adversary can not find adversarial inputs that cause the model to make mistakes. This finding initiates the desideratum to explore more efficient strategy to enforce on the model for improving Robustness, beside exploiting adversarial training framework as developed by \cite{Madry2017TowardsDL}.

We list the main contributions of our work below :
\begin{itemize}
    \item We design a model which progressively generalize a linear classifier, enforced via our regularisation method \textbf{Linearity Constraint Regularization}.
    
    \item Our trained model achieve \textbf{state of the art adversarial accuracy} compared to other adversarial training approaches which aims to improve robustness against PGD attack on MNIST, CIFAR10 and SVHN datasets.
    
    \item We also propose a two step iterative technique for \textbf{generating adversarial image} by leveraging \textbf{Inverse Representation Learning} and \textbf{Linearity} aspect of our adversarially trained model. We show that our proposed adversarial image generation method is well generalized and can act as adversary to other deep neural network classifiers also.

\end{itemize}

\section{Background and Related Work}
\label{sec:2}

In this section, we briefly discuss about adversarial attack problem formulation in classification framework and then describe Projected Gradient Descent (PGD) \cite{Madry2017TowardsDL} adversarial attack which we use as a baseline in our paper.

The goal of an adversarial attack is to find out minimum perturbation $\delta$ in the input space $x$ (i.e. input pixels for an image) that results in the change of class prediction by the network $\psi$. Mathematically, we can write this as follows:
\begin{gather*}
 \operatorname*{argmin}_\delta\>|| \delta ||_{p\text{-}norm} \> \>\>s.t\>\> \psi_{k}(x + \delta;\theta) \geq \>\psi_{k^{\prime}}(x;\theta)
\end{gather*}
Here, $\psi_{k^{\prime}}$ and $\psi_{k}$ represents the classifier output component corresponding to the true class and any other class except the true class respectively. $\theta$ denotes the parameters of the classifier $\psi$. Input and output are represented as $x$ and $y$ respectively and the objective function as $J(\theta,x,y)$. 

The magnitude of adversarial perturbation is constrained by a $p\text{-}norm$ where $p\text{-}norm \in \{ 0,2,\infty \}$ to ensure that the adversarially perturbed example $x_{adv}$ is close to the original sample. Below, we pose this as an optimization problem:
\begin{equation}
    x_{adv}=\operatorname*{argmax}_{x^\prime} \> J(\theta,x^\prime,y) \>\>\> s.t. {|| x^\prime - x ||_{p\text{-}norm} < \epsilon}
    \label{eq:1}
\end{equation}
Projected Gradient Descent \cite{Madry2017TowardsDL} is an iterative method to generate perturbed output using Fast Gradient Sign Method (FGSM) \cite{Goodfellow2014ExplainingAH} confined to bounded perturbation space $S$ and can be formulated as follows:
\begin{equation}
    x^{i+1}\> =\> Proj_{x+S}\>(x^{i}\>+\> \alpha \sgn{(\nabla_{x} J(\theta,x^{i},y)))}
    \label{eq:2}
\end{equation}
Here $x^{i}$ represents perturbed sample at i-th iteration.

In general, there are mainly two types of adversarial attacks, white box and black box attacks. In white box attack, adversary has complete knowledge of the network and its parameters. While in Black box attack there is no information available about network architecture or parameters. Numerous works have been proposed to defend against such adversarial attacks. \cite{46641, DBLP:journals/corr/abs-1801-02613,DBLP:journals/corr/abs-1711-00117, DBLP:journals/corr/abs-1803-01442,article, DBLP:journals/corr/abs-1711-01991, DBLP:journals/corr/abs-1710-10766} showed different defense strategies to improve the robustness of deep neural networks. Unfortunately, \cite{DBLP:journals/corr/abs-1802-00420} find that all these defense policies use obfuscated gradients, a kind of gradient masking, that leads to false sense of security in defense against adversarial examples and provides a limited improvement in robustness. \cite{DBLP:journals/corr/abs-1802-00420} also observes Adversarial training is the only defense which significantly increases robustness to adversarial examples within the threat model. 

Hence, we focus on adversarial training framework which allows to defend against adversarial attack by training neural networks on adversarial images that are generated on-the-fly during training. Adversarial training constitutes the current state of the art in adversarial robustness against white-box attacks. In this work we aim to improve model robustness further by leveraging adversarial training framework.

\cite{DBLP:journals/corr/abs-1812-03411} proposed a network architecture (termed as FDT) that comprises of denoising blocks at the intermediate hidden layers which aims to remove noise from the features in latent layers introduced due to adversarial perturbation, in turn increases the adversarial robustness of the model. \cite{DBLP:journals/corr/abs-1905-05186} observes that latent layers of an adversarially trained model is not robust and proposed a technique to perform adversarial training of latent layer (termed as LAT) in conjunction with adversarial training of the full network which aims at increasing the adversarial robustness of the model. In this paper we have shown improvements in adversarial accuracy as well as clean accuracy over these aforementioned methods.

\section{Neural Network Architecture and Training Objective Function}
\label{sec:3}

\begin{figure*}
  \includegraphics[width=\textwidth,height=8.8cm]{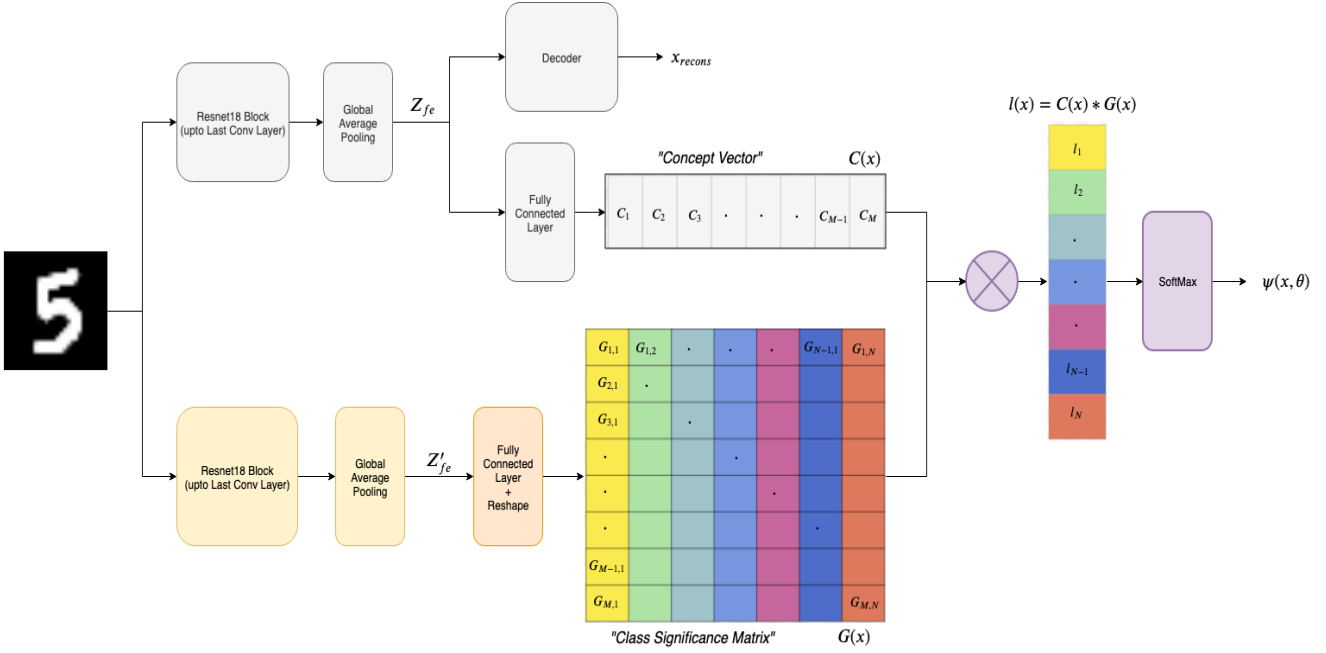}
  \caption{Network Architecture of our proposed neural network classifier LiCS Net.}
  \label{fig:1}
\end{figure*}

In this section, we briefly discuss our proposed network architecture and training objective. As depicted in Fig. \ref{fig:1} , our network architecture consists of two separate branches: Concept and Significance branch. Concept branch consists of series of stacked residual convolution block of ResNet18  \cite{DBLP:journals/corr/HeZRS15} architecture along with global average pooling and two parallel components: fully connected layer and series of up-Convolution layers. Significance branch consists of series of stacked residual convolution block of ResNet18 \cite{DBLP:journals/corr/HeZRS15} architecture along with global average pooling and fully connected combined with reshape layer.

$x_{0}$ is fed as input to Concept branch to output \enquote{Concept Vector} $C(x)$ from fully connected layer and reconstructed image $x_{recons}$ from series of up-convolution layer. $x_{0}$ is also fed as input to Significance branch to produce \enquote{Class Significance Matrix} $G(x)$, a 2-Dimensional Matrix as output. Finally, we multiply \enquote{Concept Vector} $C(x)$ with \enquote{Class Significance Matrix} $G(x)$, which forms the logit vector $l(x)$ and apply the softmax function to produce class probabilities of the classifier $\psi$.

Let's assume, Concept branch of our network is defined by a sequence of transformations $h_{k}$ for each of its $K$ layers. For an input $x_{0}$ (which we represent as $C_{0}$), we can formulate the operation performed by the Concept branch of our architecture as below:
\begin{equation}
    C_{K}\>=\>h_{k}(C_{k-1})\>\>\> \text{ for } \> k\>=\>1,...,K
    \label{eq:3}
\end{equation}
Here \enquote{Concept Vector} $C_{K} \in\>\mathbb{R}^{M}$, where $M$ represents the number of different concepts, we want to capture from underlying data distribution.

Similarly we assume, Significance branch of the network is defined by a sequence of transformations $f_{k}$ for each of its $K$ layers. We can formulate the operation performed by the Significance branch of our architecture with same input $x_{0}$ (which we represent as $G_{0}$) as below:
\begin{equation}
    G_{K}\>=\>f_{k}(G_{k-1})\>\>\>\text{ for }\> k\>=\>1,...,K
    \label{eq:4}
\end{equation}
Here \enquote{Class Significance Matrix} $G_{K}\in\>\mathbb{R}^{M*N}$, where, $N$ and $M$ represents number of classes and concepts respectively. 

We define logit vector $l$ as product of Concept vector $C_{K}$ and Class Significance Matrix $G_{K}$. We formulate the logit vector $l$ and output of classifier $\psi$ as follows:
\begin{equation}
    l(x,\theta) = C_{K}(x,\theta) \cdot G_{K}(x,\theta)
    \label{eq:5}
\end{equation}
\begin{equation}
    \psi(x,\theta) = Softmax\>(l(x,\theta))
    \label{eq:6}
\end{equation}

We term the above proposed network architecture as \textbf{Linearised Concept Significance Net} denoted as \textbf{LiCS} \textbf{Net}. In the next section, We shed lights to explain linear aspect in our classifier.

We also propose a training objective termed as \textbf{Lipschitz and Linearity Constrained Objective Function} as \textbf{Obj\textsubscript{{L\textsuperscript{2}C}}}. It comprises of four different components : \textbf{(a)} Cross-Entropy Loss for Classification, denoted as $\mathcal{L}_{CE}$, \textbf{(b)} Linearity Constraint Regularization loss, which is denoted as $\mathcal{L}_{LC}$, \textbf{(c)} Local Lipschitz Regularizer for Concept Vector and Class Significance Matrix, which is denoted as $\mathcal{L}_{LR}$ and \textbf{(d)} Reconstruction error, which is represented as $\mathcal{L}_{RE}$. Our training objective $Obj_{L^{2}C}$ can be formulated as follows:
\begin{multline}
    Obj_{L^{2}C}(x,x_{adv},\theta,y)\>=\>\alpha \cdot \mathcal{L}_{CE}\>+\>\>\beta \cdot \mathcal{L}_{LC}\>\\+\>\gamma \cdot \mathcal{L}_{LR}\>+\>\lambda \cdot \mathcal{L}_{RE}
    \label{eq:7}
\end{multline}
Here $\alpha$, $\beta$, $\gamma$ and $\lambda$ are the regularizer coefficient corresponding to different component of our objective function.

We briefly discuss about different component of the objective function.

(a) \textbf{Cross-Entropy loss} is defined as follows:
\begin{equation}
\begin{split}
    \mathcal{L}_{CE} = \sum_{c=1}^{N} (y_{true} \cdot log(y_{pred}))\\
    \>\>where\>\>y_{pred}=\psi(x_{adv},\theta)
    \label{eq:8}
\end{split}
\end{equation}
Here $c$ represents class index, $N$ denotes number of classes, $y_{true}$ holds ground-truth class label score. $y_{pred}$ hold class probability scores when an adversarial image $x_{adv}$ is being fed through the classifier. 

For the remainder of this section, $x_{adv}$ refers to the generated adversarial image from its corresponding true image $x$ by following eq. \ref{eq:2}. Note that ground-truth label is same for both $x_{adv}$ and $x$.

(b) \textbf{Linearity Constraint Regularization loss} is defined as follows:
\begin{equation}
    \mathcal{L}_{LC}\>=\> \norm{\frac{\partial l(x_{adv},\theta)}{\partial C_{K}(x_{adv},\theta)}\>-\>G_{K}(x_{adv},\theta)}_{2}
    \label{eq:9}
\end{equation}
Here $||\> .\> ||_{2}$ represents the $l2\text{-}norm$ of a given vector. We are enforcing the logit vector $l(x_{adv},\theta)$ to behave as a linear classifier with \enquote{Concept Vector} $C_{K}$ as the input and \enquote{Class Significance Matrix} $G_{K}$ as the parameters or weights of that linear classifier. 

Intuitively, a linear classifier can be defined as $L(x,W) = W^{T} \cdot x$, where $x$ is input and $W$ captures significance of each input features for the classification task. Comparing eq. \ref{eq:5} with linear classifier $L$, we hypothesize $C_{K}$ encodes the transformed input features like $x$ and $G_{K}$ encodes the significance weight factor for concept features same as $W$.

(c) \textbf{Local Lipschitz Regularizer for Concept Vector and Class Significance Matrix} is defined based on the concept of \enquote{Lipschitz constant}. Let's assume a function $\phi(x)$ having Lipschitz Constant value of $L_{\Phi}$. Below equation holds true for all $\Delta$.
\begin{equation}
    \norm{\phi(x+\Delta)\>-\>\phi(x) } \leq\>L_{\Phi} \cdot \norm{\Delta}
    \label{eq:10}
\end{equation}

Hence, we argue that a lower Lipschitz constant ensures that the function’s output for its corresponding perturbed input is not significantly different and it can be used to improve the adversarial robustness of a classifier. 

The Lipschitz constant of our network $\psi$ has an upper bound defined by the product of Lipschitz constant of its sub-networks: Concept and Significance branch. We can formulate this as follows:
\begin{equation}
    L_{\psi}\>\leq\> L_{sub_{C}}\> \cdot \>L_{sub_{G}}
    \label{eq:11}
\end{equation}
Here $L_{sub_{C}}$ and $L_{sub_{G}}$ are the Lipschitz constant of the sub network Concept branch and Significance branch respectively.

To achieve a robust classifier, we aim to improve robustness of its sub-networks using Local Lipschitz Regularizer for Concept Vector and Class Significance Matrix, which is formulated as follows:
\begin{equation}
 \begin{aligned}
    \mathcal{L}_{LR}\>=\>\beta_{1} \cdot || C_{K}(x,\theta) - C_{K}(x_{adv},\theta)||_{2} \>\\ +\beta_{2} \cdot || G_{K}(x,\theta) - G_{K}(x_{adv},\theta)||_{2}
 \end{aligned}
 \label{eq:12}
\end{equation}
Here, $C_{K}(x,\theta)$ and $G_{K}(x,\theta)$ are Concept vector and Class Significance Matrix respectively when a true image is passed through our network. Similarly, $C_{K}(x_{adv},\theta)$ and $G_{K}(x_{adv},\theta)$ are Concept vector and Class Significance Matrix respectively when the corresponding adversarial image is passed through the network. $\beta_{1}$ and $\beta_{2}$ are the regularization coefficients. 

Note that, we enforce invariability of Concept vector $C_{K}$ and Class Significance matrix $G_{K}$ to the model for an original image $x$ and its corresponding adversarial image $x_{adv}$ through Local Lipschitz Regularizer.

(d) \textbf{Reconstruction error} is defined as follows:
\begin{equation}
    \mathcal{L}_{RE}\>= || x\>-x_{recons}||_{2}
    \label{eq:13}
\end{equation}
where $x$ is the input image and $x_{recons}$ is the reconstructed image at last up-convolution layer of the Concept branch of our network.


\section{Adversarial Image Generation}
\label{sec:4}
In this section, we propose adversarial image generation method which consists of two iterative steps stated as follows: \textbf{(i)} Find minimum perturbation of Concept Vector $C(x,\theta)$ which guarantees to fool the classifier by exploiting \enquote{linearity} aspect of the classifier. \textbf{(ii)} Using the perturbed concept vector from step (i), generate an equivalent perturbation in image space (i.e. image pixels) by leveraging inverse representation which fools the classifier.

We elaborate above steps of adversarial image generation method below:

\textbf{Step (i)}: We represent the i-th component of logit vector defined in eq. \ref{eq:5} as:
\begin{equation}
    l^{i}(x,\theta)\>=\>C_{K}(x,\theta)\> \cdot \>G^{i}_{K}(x,\theta)
    \label{eq:14}
\end{equation}
Here $G^{i}_{K}(x,\theta)$ is the i-th column of Class Significance Matrix which encodes the significance weight factor of concepts for the i-th class.

Let \textquotesingle s assume for a given image $x_{0}$, the network produces a Concept Vector $C_{K}$, and the ground truth class label corresponding to image $x_{0}$ is $\Tilde{i}$. Now, the minimum perturbation ($\delta$) required in Concept Vector to fool the classifier can be formulated as follows:

\begin{gather*}
    \operatorname*{argmin}_{\delta} || \delta ||_{2} \>\>\>s.t.
\end{gather*}
\begin{math}
\begin{aligned}
    (C_{K_{0}}(x_{0},\theta)+\delta) \cdot G^{i}_{K}(x_{0},\theta)\>\geq\>(C_{K_{0}}(x_{0},\theta)+\delta)\\
    \textbf{        } \cdot G^{\Tilde{i}}_{K}(x_{0},\theta) \nonumber
\end{aligned}
\end{math}

In the above relation, \enquote{$i$} represents the class index and $i \neq \Tilde{i}$ must satisfy. Now, we can re-arrange and write above equation as follows:
\begin{equation}
    \delta\>\leq\>\frac{C_{K}(x_{0},\theta) \cdot (G^{i}_{K}(x_{0},\theta)-G^{\Tilde{i}}_{K}(x_{0},\theta))}{G^{\Tilde{i}}_{K}(x_{0},\theta)-G^{i}_{K}(x_{0},\theta)}
    \label{eq:15}
\end{equation}

Using eq. \ref{eq:5} into eq. \ref{eq:15}, we write this as follows:
\begin{equation}
    \delta\>\leq\>\frac{l^{i}(x_{0},\theta)-l^{\Tilde{i}}(x_{0},\theta)}{G^{\Tilde{i}}_{K}(x_{0},\theta)-G^{i}_{K}(x_{0},\theta)}
    \label{eq:16}
\end{equation}

\begin{figure}[htp]
    \centering
    \includegraphics[width=8cm]{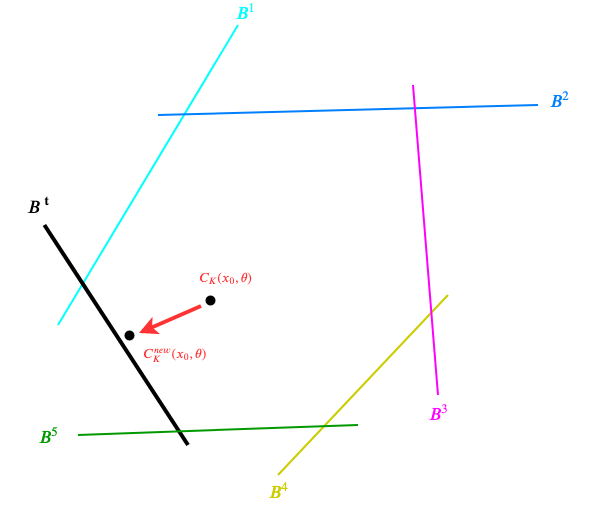}
    \caption{depicts the optimal shift of Concept Vector towards the normal direction of the nearest class decision boundary plane in order to change the classifier prediction.
    }
    \label{fig:2}
\end{figure}

As shown in Fig. \ref{fig:2}, Concept vector is represented as a point in $M$ dimensional concept space. We decide to perturb the Concept vector $C_{K}(x_{0},\theta)$ towards the linear decision boundary of a target class $i^{t}$, which satisfy the following relation:
\begin{equation}
    i^{t}\>=\>\operatorname*{argmin}_{i\neq\Tilde{i}}\frac{|l^{i}(x_{0},\theta)-l^{\Tilde{i}}(x_{0},\theta)|}{||G^{\Tilde{i}}_{K}(x_{0},\theta)-G^{i}_{K}(x_{0},\theta)||_{2}}
    \label{eq:17}
\end{equation}
Here, $i^{t}$ is the target class index, which has the decision boundary nearest to the Concept vector $C_{K}(x_{0},\theta)$ compared to decision boundary of any other class in terms of $||\>.\>||_{2}$ norm. 

We then shift the Concept vector $C_{K}(x_{0},\theta)$ in the direction normal to the $i^{t}$ linear decision boundary surface. As shown in Fig. \ref{fig:2}, the movement of Concept vector is represented by a red arrow and the linear decision boundary of the target class $i^{t}$ is depicted as $B^{t}$. Mathematically, we represent the direction of the optimal shift ($\delta^{*}$) of the concept vector $C_{K}(x_{0},\theta)$ as follows:
\begin{equation}
    \delta^{*}\>= \frac{|l^{i^{t}}(x_{0},\theta)-l^{\Tilde{i}}(x_{0},\theta)}{||G^{i^{t}}_{K}(x_{0},\theta)-G^{\Tilde{i}}_{K}(x_{0},\theta)||_{2}^{2}}\> \cdot \>(\nabla (l^{i^{t}}) - \nabla (l^{\Tilde{i}}))
    \label{eq:18}
\end{equation}
Hence, we obtain a new perturbed concept vector $C^{new}_{K}(x_{0},\theta)$, by adding the optimal shift to the actual value of concept vector $C_{K}(x_{0},\theta)$. We represents this as follows:
\begin{equation}
    C^{new}_{K}(x_{0},\theta)\>=\>C_{K}(x_{0},\theta) +\delta^{*}
    \label{eq:19}
\end{equation}

We perform this whole process iteratively, until the perturbed concept vector change the classifier's true prediction. We denote the final perturbed concept vector as $C^{perturb}_{K}(x_{0},\theta)$ which obeys below condition:
\begin{equation}
    \begin{gathered}
        C^{perturb}_{K_{0}}(x_{0},\theta) \cdot G^{i}_{K}(x_{0},\theta)\geq C_{K_{0}}(x_{0},\theta) \cdot G^{\Tilde{i}}_{K}(x_{0},\theta)\\
        such\>that \>\>\> i\>\neq\>\Tilde{i}
    \end{gathered}
    \label{eq:20}
\end{equation}

\textbf{Step (ii)} : From step (i), we obtain a final perturbed concept vector $C^{perturb}_{K}(x_{0},\theta)$ which guarantees to change the classifier's prediction. Now the main question is \enquote{How do we generate the adversarial image ($x_{adv}^{perturb}$) which corresponds to the final Perturbed Concept vector $C^{perturb}_{K}(x_{0},\theta)$?}. 

As illustrated in \cite{DBLP:journals/corr/abs-1906-00945}, the Representation vector of a given input $x$ are the activations of the penultimate layer of the network. \cite{DBLP:journals/corr/abs-1906-00945} also illustrate that adversarially trained model produces similar Representation vectors for semantically similar images. For a given input, Concept vector generated from the Concept branch of our network LiCS Net is analogous to Representation vector.

We have used \enquote{Inverse Representation} \cite{DBLP:journals/corr/MahendranV14,DBLP:journals/corr/abs-1711-10925,DBLP:journals/corr/abs-1906-00945} method to generate adversarial image $x_{adv}^{perturb}$ which corresponds to the final Perturbed Concept vector $C^{perturb}_{K}(x_{0},\theta)$,  starting from the true input image $x_{0}$ and its corresponding Concept vector $C_{K}(x_{0},\theta)$. We can pose this method as an optimization problem detailed in Algorithm 1.

\begin{algorithm}
\KwIn{True Image $x_{0}$, Perturbed Concept Vector $C_{K}^{perturb}(x_{0},\theta)$, LiCS Net Classifier $\psi_{LiCS}$}
\KwOut{Adversarial Image $x_{adv}^{perturb}$}

Initialise $x_{adv}$ = $x_{0}$\\
Set $C_{K}^{target} = C_{K}^{perturb}(x_{0},\theta)$\\
\textbf{while} 
\linebreak
{$\operatorname*{argmax}\> \psi_{LiCS}(x_{adv}, \theta) \neq  \operatorname*{argmax}\> \psi_{LiCS}(x_{0}, \theta)$ }
\linebreak
\textbf{do}
\linebreak
(i) $x_{adv} = x_{adv} - \nabla_{x_{adv}} ||C_{K}^{target} - C_{K}(x_{adv}, \theta)||$
\linebreak
(ii) if $\norm{x_{adv} - x_{0}}_{2\text{-}norm} \geq \epsilon$ 
\linebreak
$\textbf{     } \>\>\>\>\> then,\>\> x_{adv} = Proj_{x_{0} + S}(x_{adv})$

Adversarial Image, $x_{adv}^{perturb} = x_{adv}$

\caption{Pseudo-code of Step (ii)}
\label{algo:b}
\end{algorithm}

We term above proposed algorithm as \textbf{Adversarial Image Generation using Linearity and Inverse Representation} as \textbf{AIG\textsubscript{LiIR}}. 

\begin{figure*}
  \includegraphics[width=\textwidth,height=8cm]{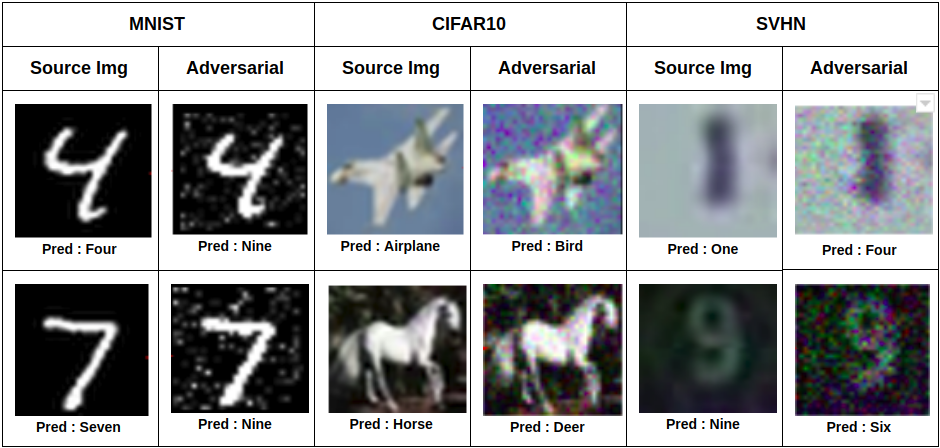}
  \caption{depicts few sample pairs of source image and its corresponding adversarial image generated by our proposed $AIG_{LiIR}$ method.}
  \label{fig:3}
\end{figure*}

In Fig. \ref{fig:3}, we depicted few adversarial images generated by our proposed method $AIG_{LiIR}$. Such adversarial image guarantees to fool white box threat model. For example, in the case of CIFAR10, generated adversarial images by $AIG_{LiIR}$ method are being classified by our proposed LiCS Net as \textit{Bird} and \textit{Deer} whereas its corresponding true level are \textit{Airplane} and \textit{Horse}.

Moreover with evident results provided in next section, we claim that our adversarial image generation method $AIG_{LiIR}$ is very effective in fooling various deep neural network designed for classification task.

We infer most deep neural network classifier consists of series of feature extraction layers or convolutional layers followed by one or more fully connected layers. Hence, we represents such deep neural network classifier $\Psi$ as: $\Psi(x,\theta) = g(f(x))$. Here, $f(x)$ act as the transformation applied on input by sub network of $\Psi$ which comprises of all the layers upto penultimate layer and $g$ act as the significance factor applied by last fully connected layer, to each transformed features from $f(x)$.

We observe that f and g are analogous to Concept vector and Significance Matrix of our network LiCS Net such that Concept vector encodes the transformed features of input image and Significance Matrix captures significance weight factor for each concept features. We argue that our classifier LiCS Net is linear on its transformed input features. Hence we can also apply our adversarial image generation $AIG_{LiIR}$ to other classification network.

\section{Experiments and Results}
\label{sec:5}
In this section, we briefly discuss our experimental findings to analyze the robustness of our proposed network LiCS Net. We have performed experiment to show the \enquote{Transferability} of the adversarial images generated by our method AIG\textsubscript{LiIR}. We also discuss findings on the effectiveness of Linearity Constraint  and Local Lipschitz Regularizer in our objective function $Obj_{L^{2}C}$ to improve adversarial robustness.
The samples are constructed using PGD adversarial perturbations as proposed in \cite{Madry2017TowardsDL} for adversarial training. For the remainder of this section, we refer an adversarially trained model as a model which is trained using PGD adversarial examples \cite{Madry2017TowardsDL}. Clean and Adversarial accuracy are defined as a accuracy of a network over original images and over the adversarial images generated from the test dataset respectively. Higher adversarial accuracy implies more robust network. To test the efficacy of our proposed training method, we perform experiments over CIFAR10, SVHN and MNIST dataset. For fairness, we compare our method against three different fine-tuned techniques applied in adversarial training framework. These are Adversarial Training (AT) \cite{Madry2017TowardsDL}, Feature Denoising Training (FDT) \cite{DBLP:journals/corr/abs-1812-03411} and Latent adversarial Training( (LAT) \cite{DBLP:journals/corr/abs-1905-05186}.

\subsection{\textbf{Evaluation using Adversarial Accuracy}}

We train LiCS Net using $Obj_{L^{2}C}$ training objective and denote this as Concept-Significance Adversarial Training (CSAT). In order to compare Concept-Significance Adversarial Training (CSAT) with other fine tuned adversarial techniques we use PGD adversarial perturbation.

\textbf{PGD configuration }:
The configuration of PGD adversarial perturbation varies for different datasets. For MNIST dataset, we restrict the maximum amount of per pixel perturbation as $0.3$ (in the pixel scale of 0 to 1 range) and we choose 40 steps of PGD iterations with step size of 2/255. For CIFAR10 and SVHN dataset, we restrict the maximum amount of per pixel perturbation as $8.0$ (in the pixel scale of 0 to 255 range) and we choose 10 steps of PGD iterations with step size of 2/255.

\textbf{Concept-Significance Adversarial Training(CSAT) vs others }:
Table \ref{tab:1} reports the adversarial accuracy comparison between our proposed technique Concept-Significance Adversarial Training (CSAT) and  other fine tuned adversarial training techniques such as Adversarial Training (AT), Feature Denoising Training (FDT), Latent adversarial Training(LAT) over different datasets. Concept-Significance Adversarial Training (CSAT) outperforms all the other fine tuned adversarial training strategies such as AT, FDT, LAT and achieves state of the art adversarial accuracy on various datasets such as MNIST, CIFAR10 and SVHN.

Concept-Significance Adversarial Training (CSAT) achieves adversarial accuracy of 54.77\%, 60.41\% and 98.68\% on CIFAR10, SVHN and MNIST dataset respectively. Note that in AT, FDT and LAT, fine tuning strategies are being performed over Wide ResNet based (WRN 32-10 wide) architecture \cite{BMVC2016_87} in the case of CIFAR10 and SVHN dataset. We use the same network configuration for all the different datasets. Note that we choose dimension of concept vector as 10 and the dimension of class significance matrix as 10$\times$10. To train our model we use adam optimizer \cite{Kingma2014AdamAM} with the learning rate of 0.0002.

\begin{table}[htbp]
	\begin{center}
		\begin{tabular}{|p{1.30cm}|p{1.9cm}|p{1.75cm}|p{1.45cm}|}
		\hline
			\textbf{Dataset} & \textbf{Adversarial Training Techniques}& \textbf{Adversarial Accuracy}&
			\textbf{Clean Accuracy} \\
			\hline
            \multirow{3}{*}{\textbf{CIFAR10}}
            & AT & 47.12\% & 87.27\% \\
            & FDT & 46.99\% & 87.31\% \\ 
            & LAT & 53.84\% & 87.80\% \\ 
            & \textbf{CSAT} & \textbf{54.77\%} & \textbf{87.65\%} \\
            \hline
            \multirow{3}{*}{\textbf{SVHN}}
            & AT & 54.58\% & 91.88\% \\
            & FDT & 54.69\% & 92.45\% \\ 
            & LAT & 60.23\% & 91.65\% \\ 
            & \textbf{CSAT} & \textbf{61.37\%} & \textbf{95.11\%} \\
            \hline
            \multirow{3}{*}{\textbf{MNIST}}
            & AT & 93.75\% & 98.40\% \\
            & FDT & 93.59\% & 98.28\% \\ 
            & LAT & 94.21\% & 98.38\% \\ 
            & \textbf{CSAT} & \textbf{98.68\%} & \textbf{99.46\%} \\
            \hline
		\end{tabular}
		\label{tab1}
	\end{center}
		\caption{Comparison of Adversarial accuracy and Clean accuracy for CIFAR10, SVHN and MNIST datasets acheieved by different adversarial training techniques}
		\label{tab:1}
\end{table}

\subsection{\textbf{Transferability of Generated Adversarial Samples}}

Training on adversarial samples has been shown as one of the best method to improve robustness of a classifier \cite{Madry2017TowardsDL,Tramr2017EnsembleAT,DBLP:journals/corr/abs-1905-05186}. We denote $AIG_{CSPGD}$ as the adversarial image generation method to generate adversarial samples using PGD adversarial perturbations \cite{Madry2017TowardsDL} on our trained network LiCS Net using our objective function $Obj_{L\textsuperscript{2}C}$. In previous section, we proposed our adversarial image generation method $AIG_{LiIR}$ to generate adversarial samples using our trained network LiCS Net.

We evaluate adversarial accuracy of different standard pre-trained models such as GoogleNet \cite{DBLP:journals/corr/SzegedyLJSRAEVR14} (trained on CIFAR10), ResNet34 \cite{DBLP:journals/corr/HeZRS15} (trained on SVHN) and custom Net-3FC (trained on MNIST) on adversarial samples generated using $AIG_{LiIR}$ and $AIG_{CSPGD}$ method and compare with their respective clean test accuracy. Note that custom Net-3FC consists of 3 fully connected layers along with Relu activation and softmax function.

\begin{figure}[htp]
    \centering
    \includegraphics[width=8.3cm]{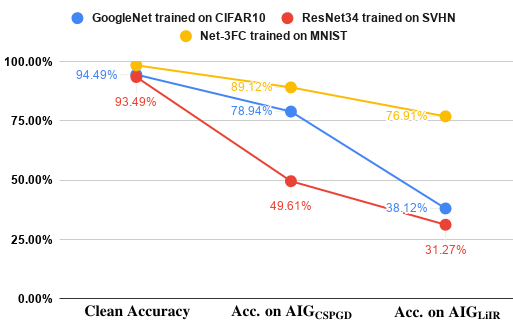}
    \caption{ depicts the comparison between clean accuracy and adversarial accuracy evaluated on samples generated by $AIG_{CSPGD}$ and $AIG_{LiLR}$ method.}
    \label{fig:4}
\end{figure}


\textbf{ AIG\textsubscript{LiIR} vs AIG\textsubscript{CSPGD} }: As evident in Fig. \ref{fig:4}, For classifiers GoogleNet, ResNet34 and custom Net-3FC, we observe significant drop in accuracy on adversarial samples generated using $AIG_{CSPGD}$ and $AIG_{LiIR}$ method, in comparison to their clean accuracy. For example in case of ResNet34, drop in accuracy on adversarial samples generated using $AIG_{LiIR}$ method is 18.34\% more than drop in accuracy on adversarial samples generated using $AIG_{CSPGD}$ in comparison with clean accuracy on SVHN dataset. Similary in case of GoogleNet, drop in accuracy on adversarial samples generated using $AIG_{LiIR}$ method is 40.82\% more than drop in accuracy on adversarial samples generated using $AIG_{CSPGD}$, in comparison with clean accuracy on CIFAR10 dataset.

These experimental finding suggests adversarial samples generated using $AIG_{LiIR}$ act as a better adversary to various network architectures such as GoogleNet, Resnet34 and custom Net-3FC which shows the \enquote{transferability} of these adversarial samples. Our adversarial image generation method $AIG_{LiIR}$ can also be used for black box attack to other classification networks due to its robustness and transferability. 

\begin{figure}[htp]
    \centering
    \includegraphics[width=8.3cm]{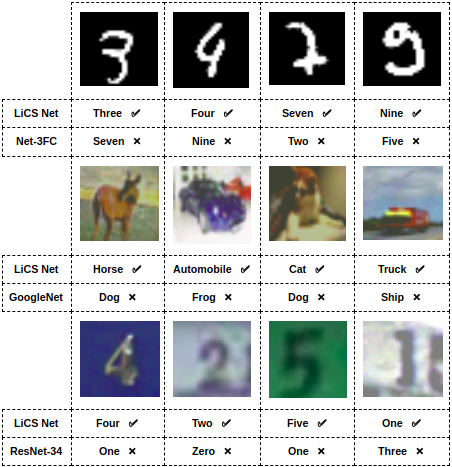}
    \caption{ depicts few samples of adversarial image generated by $AIG_{CSPGD}$ method, which can fool standard classifiers such as GoogleNet, ResNet34 and custom Net-3FC trained respectively on CIFAR10, SVHN and MNIST datasets but LiCS Net predicts the class correctly for all these adversarial images.}
    \label{fig:5}
\end{figure}

In Fig. \ref{fig:5}, we depict few adversarial samples generated by $AIG_{CSPGD}$ method using the original test set images of the MNIST, CIFAR10 and SVHN datasets. These generated samples fooled standard pre-trained networks such as custom Net-3FC, GoogleNet and ResNet34. But our proposed model LiCS Net, trained using our proposed objective function $Obj_{L^{2}C}$ as described in eq. \ref{eq:7} correctly predicts the true class of these generated adversarial samples.


\subsection{\textbf{Ablation Study}}
 
\textbf{Architecture} : We have used our proposed network LiCS Net for ablation experiments on our objective function as stated in eq. \ref{eq:7}. We kept all the hyperparameters same for different experiments of ablation study. We evaluated adversarial accuracy of our network LiCS Net trained with base objective function (termed as $Obj_{BASE}$) which consists of cross-entropy and reconstruction loss and used it as our baseline adversarial accuracy for ablation study.

\textbf{Linearity Constraint vs Local Lipschitz Regularizer} :
As evident from Table \ref{tab:2}, adversarial accuracy improves in both the cases when base training objective is augmented with Linearity Constraint Regularizer (termed as $Obj_{BASE+LCR}$) and also when $Obj_{BASE}$ is augmented with Local Lipschitz Regularizer (termed as $Obj_{BASE+LLR}$) compared to the baseline adversarial accuracy. Experimental findings suggests that the regularisation impact of Linearity Constraint is much more effective compared to Local Lipschitz Regularisation to improve model robustness. Note that we achieve state of the art adversarial accuracy when $Obj_{BASE}$ is augmented with both Linearity Constraint and Local Lipschitz Regulariser.
\begin{table}[htbp]
	\begin{center}
		\begin{tabular}{|p{2.20cm}|p{1.55cm}|p{1.40cm}|p{1.35cm}|}
		\hline
			\multirow{2}{*}{\textbf{Training Obj.}} & \multicolumn{3}{c|}{\textbf{Adv. Accuracy}}\\
			\cline{2-4}
			& \textbf{CIFAR10}& \textbf{SVHN} & \textbf{MNIST} \\
			\hline
            {\textbf{$Obj_{\textbf{BASE}}$}}
            & 47.02\% & 54.17\% & 93.14\% \\
            \hline
            {\textbf{$Obj_{\textbf{BASE+LLR}}$}}
            & 49.75\% & 55.70\% & 95.39\% \\
            \hline
            {\textbf{$Obj_{\textbf{BASE+LCR}}$}}
            & 54.39\% & 60.61\% & 98.06\% \\
            \hline
            {\textbf{Obj\textsubscript{{L\textsuperscript{2}C}} }}
            & \textbf{54.77}\% & \textbf{61.37}\% & \textbf{98.68}\% \\
            \hline
		\end{tabular}
	\end{center}
		\caption{{depicts adversarial accuracy comparison of our model LiCS Net trained using different objective functions on various datasets}}
		\label{tab:2}
\end{table}	

\section{Conclusion}
We observe that adversarial training method is the de-facto method to improve model robustness. We propose the model \textbf{LiCS Net} which achieves state of the art adversarial accuracy trained with our proposed objective function on the MNIST, CIFAR10 and SVHN datasets along with the improvement in normal accuracy. We also propose an Adversarial Image Generation method  \textbf{AIG\textsubscript{LiIR}} that exploits Linearity Constraint and Inverse Representation learning to construct adversarial examples. We performed several experiments to exhibit the robustness and transferability of the adversarial samples generated by our Adversarial Image Generation method across different networks. We demonstrate that the model trained with Linearity Constrained Regularizer in the adversarial training framework boost adversarial robustness. Through our research, we shed lights on the impact of linearity on robustness. We hope, our findings will inspire discovery of new adversarial defenses and attacks and, offers a significant pathway for new developments in adversarial machine learning.

{\small
\bibliographystyle{ieee}
\bibliography{egbib}
}

\end{document}